# Continual Reinforcement Learning for Cyber-Physical Systems: Lessons Learned and Open Challenges

**Kim N. Nolle**
School of Computer Science and Statistics
Trinity College Dublin
Dublin, Ireland
nollek@tcd.ie

**Ivana Dusparic**
School of Computer Science and Statistics
Trinity College Dublin
Dublin, Ireland
ivana.dusparic@tcd.ie

**Rhodri Cusack**
Trinity College Institute of Neuroscience
Trinity College Dublin
Dublin, Ireland
cusackrh@tcd.ie

**Vinny Cahill**
School of Computer Science and Statistics
Trinity College Dublin
Dublin, Ireland
vjcahill@tcd.ie

## Abstract

Continual learning (CL) is a branch of machine learning that aims to enable agents to adapt and generalise previously learned abilities so that these can be reapplied to new tasks or environments. This is particularly useful in multi-task settings or in non-stationary environments, where the dynamics can change over time. This is particularly relevant in cyber-physical systems such as autonomous driving. However, despite recent advances in CL, successfully applying it to reinforcement learning (RL) is still an open problem.

This paper highlights open challenges in continual RL (CRL) based on experiments in an autonomous driving environment. In this environment, the agent must learn to successfully park in four different scenarios corresponding to parking spaces oriented at varying angles. The agent is successively trained in these four scenarios one after another, representing a CL environment, using Proximal Policy Optimisation (PPO). These experiments exposed a number of open challenges in CRL: finding suitable abstractions of the environment, oversensitivity to hyperparameters, catastrophic forgetting, and efficient use of neural network capacity.

Based on these identified challenges, we present open research questions that are important to be addressed for creating robust CRL systems. In addition, the identified challenges call into question the suitability of neural networks for CL. We also identify the need for interdisciplinary research, in particular between computer science and neuroscience.

**Acknowledgements**

This publication has emanated from research conducted with the financial support of Taighde Éireann – Research Ireland under Grant number 21/FFP-A/8957.

# 1 Introduction

There have been many advances in reinforcement learning (RL) in recent years, resulting in RL agents that are able to outperform human experts in complex tasks such as playing chess and Go [1]. However, RL agents are usually trained on a narrow set of tasks and often struggle to adapt to new tasks. This poses a problem in multi-task settings or in non-stationary environments, where the environment dynamics can change over time. This is particularly relevant to cyber-physical applications because these interact with the real world and thus have higher requirements regarding stability and robustness against uncertainty in the environment [2]. In addition, transferring agents from simulated environments to the physical environment is difficult due to the gap between simulation and reality [3].

Continual learning (CL) poses a solution to these problems. CL is a branch of machine learning that aims to improve the adaptation and generalisation capabilities of models and agents [4]. This will enable them to reapply previously learned abilities to new tasks or environments so that they can develop increasingly complex behaviours over time. Despite advancements in continual *reinforcement* learning (CRL) (see [4] for an in-depth review), there are still limitations to proposed approaches. This paper highlights open challenges in CRL based on experiments in an autonomous driving environment. In these experiments, an RL agent is trained successively to park in different parking scenarios. The results led to the identification of a set of open challenges that will inform our future work in advancing CRL. Overall, the identified challenges can be traced back to fundamental limitations of neural networks. This calls into question the suitability of neural networks as a basis for CRL. To address these challenges we highlight the need for interdisciplinary research as there is opportunity to gain inspiration from neuroscience.

# 2 Motivating Experiments

A simplified autonomous driving environment was used to explore CRL. The parking environment is an episodic goal-based environment, where the agent should park the vehicle in a specified parking space (the goal). There are three possible outcomes of an episode: the agent successfully parks in the goal parking space, the agent crashes, or the agent fails to complete the task within the maximum allowed number of time steps. Specifically, the RL agent should learn to park in four different scenarios: perpendicular parking, 25° diagonal parking, 50° diagonal parking, and parallel parking. These different parking scenarios vary only in the angle of the parking space (see Figure 1). Fundamentally, these tasks should be very similar and so it was expected that an RL agent would be able to learn to park in all four scenarios easily.

Proximal Policy Optimisation (PPO) [5], a state-of-the-art on-policy RL algorithm, was selected to train the agent. On-policy algorithms have low variance and fast convergence properties [6], suggesting that PPO should be able to learn new variations of the environment quickly and thus be suited to CL. The PPO agent is trained on the four parking scenarios successively, which represents the CL requirement. This means the same agent is trained on each scenario one after the other. The agent is trained in the following order: perpendicular parking, 25° diagonal parking, 50° diagonal parking, and parallel parking. Each scenario is trained for 500k steps. The hyperparameters of the PPO agent were tuned in the perpendicular parking scenario using random search and the same hyperparameters were used in all four scenarios.

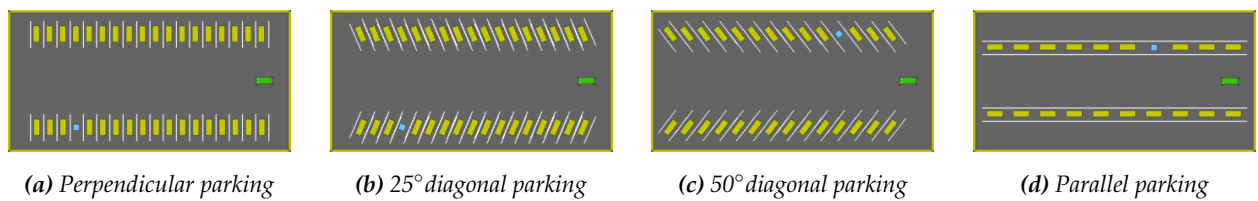

*(a) Perpendicular parking*  *(b) 25° diagonal parking*  *(c) 50° diagonal parking*  *(d) Parallel parking*

*Figure 1: The four different parking scenarios. The green rectangle represents the agent vehicle, the yellow rectangles represent other parked vehicles, and the blue square represents the goal parking space.*

Figure 2a plots the the success rate of the learned policy in the different scenarios against the timesteps during sequential training. A success rate of 1 means the episode was successful. The graph shows the average over 5 runs. It can be seen that the success rate increases in the scenario that is currently being trained. This can be seen, for example, in the perpendicular parking scenario (blue line) in the first 500k steps. However, as soon as the agent switches to learning the next scenario, the success rate decreases again. This clearly demonstrates catastrophic forgetting - the phenomenon in which previously acquired abilities are lost as new abilities are learned [7]. As the model learns a new task, the weights of the neural network are changed to produce the new desired behaviour. This in turn can result in a change in behaviour in the previously learned task.

Another observation is that, contrary to expectation, it was difficult to find a model that was capable of learning all four parking scenarios. While the agent was able to achieve 100% success rates in perpendicular and diagonal parking, it was unable to learn parallel parking (Figure 2a).

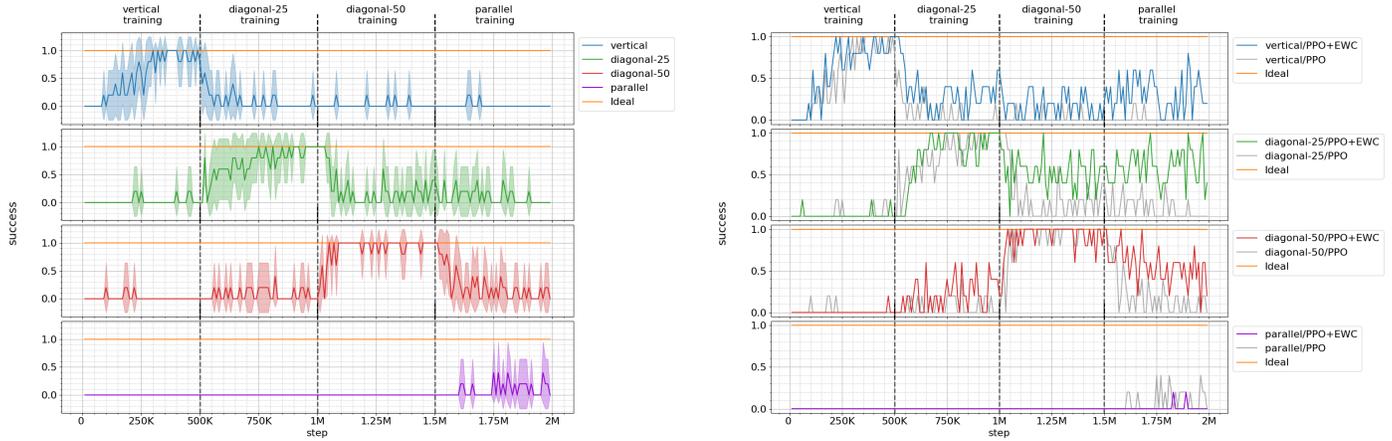

*(a) PPO agent trained sequentially in all four scenarios*    *(b) PPO+EWC agent trained sequentially in all four scenarios*

*Figure 2: Comparison of successful episodes in perpendicular (blue), 25° diagonal (green), 50° diagonal (red) and parallel (purple) parking scenarios. The line graphs represent the average over 5 runs. Shaded area represents standard deviation. In (b), gray curves represent the corresponding measurements from (a) and standard deviations omitted for readability.*

## 3 Open Challenges in Continual Reinforcement Learning

This section presents insights obtained from the aforementioned experiments. We highlight the following open challenges in CRL: finding suitable abstractions of the environment, oversensitivity to hyperparameters, catastrophic forgetting, and efficient use of neural network capacity.

### 3.1 Challenge 1: Abstracting the Environment

Our experiments showed that the reward function has a significant impact on the success of the PPO agent. We defined a reward function that facilitated successful learning in the perpendicular parking environment. This reward model was used in the experiments in Figure 2a. While perpendicular and diagonal parking were learned successfully, the results in Figure 2a show that this was not the case for parallel parking. Further experimentation showed that hyperparameter tuning in the parallel parking scenario did not solve this problem. Instead modifications to the reward model had to be made for the agent to successfully parallel park. This highlights a challenge in CL: changes in the environment or new tasks may cause previously defined reward models to no longer be suitable for successful learning.

This not only applies to the reward model but other abstractions of the environment, such as the state space or action space. New tasks may require new input features or new actions. This would mean the input or output dimensions of the neural networks would have to be adapted to accommodate these changes. The question arises, how can the environment be abstracted so that the abstractions are suitable for all future tasks that the agent should learn, including unforeseen tasks. This also poses the question of how the agent can adapt to new state or action spaces.

### 3.2 Challenge 2: Oversensitivity to Hyperparameters

Another challenge that was identified is the oversensitivity of RL, and more generally neural network training, to hyperparameters. Hyperparameter tuning was required to find a solution to perpendicular parking. While the found hyperparameters worked well for perpendicular and diagonal parking (see Figure 2a), they did not lead to successful parallel parking, even with the adjusted reward model for parallel parking. Further hyperparameter tuning in the parallel parking scenario showed that different learning rates, larger experience buffers, larger minibatch sizes and longer training times were required for the PPO agent to achieve similar performance in parallel parking as in perpendicular and diagonal parking. Similar to the previous challenge, this poses the question of how to select the hyperparameters so that all future (unknown) tasks can be solved. Alternatively, algorithms to enable CRL agents to perform autotuning at runtime are needed.

### 3.3 Challenge 3: Catastrophic Forgetting

One challenge that many CL approaches aim to solve is catastrophic forgetting. As previously mentioned, catastrophic forgetting was observed during sequential training (Figure 2a). To attempt to address this problem, we used Elastic Weight Consolidation (EWC) [8], a CL approach that minimises changes to those neural network weights that are im-portant for previously-learned tasks. The results of combining PPO and EWC can be seen in Figure 2b. While the extent of forgetting is reduced, the agent still does not retain the same performance in previous tasks as new tasks are learned. Modifications of the EWC hyperparameters showed that varying these hyperparameters did not significantly improve the effect of EWC on catastrophic forgetting. Similar observations on the efficacy of EWC in preventing catastrophic forgetting are shown, for example, in [8, 9, 10].

Other CL approaches such as [9, 11] report successful mitigation of catastrophic forgetting. However, these works usually only evaluate the approaches on supervised learning tasks such as MNIST permutations, which are comparatively simple. It is therefore unclear whether successful mitigation of catastrophic forgetting transfers to more complex and difficult domains. Indeed, other works that evaluate CL approaches in both supervised learning and RL settings tend to show that catastrophic forgetting is solved to a lesser extent in RL [8, 12]. Catastrophic forgetting in CRL therefore still remains an open issue.

### 3.4 Challenge 4: Capacity of Neural Networks

The importances of the neural network weights, which are calculated in EWC using the Fisher information matrix [8], were visualised to gain insights into the behaviour of the agent. In all four parking scenarios, the first layer of the neural network had the highest importances compared to the other hidden layers. Figure 3 shows heatmaps representing a matrix of the normalised importances in the first layer when the agent is trained on a single scenario. The columns represent the input of the layer and the rows the output. Differences in important weights can be observed between the scenarios, suggesting that training in the different scenarios individually results in different learned representations.

Interestingly, when the agent is trained sequentially on all four scenarios, there is a much stronger overlap in important weights between the scenarios (see Figure 4). This suggests that the same representations are being used for all scenarios rather than learning new representations for each scenario. The weights learned during sequential training correspond to the weights important to perpendicular parking (Figure 3a).

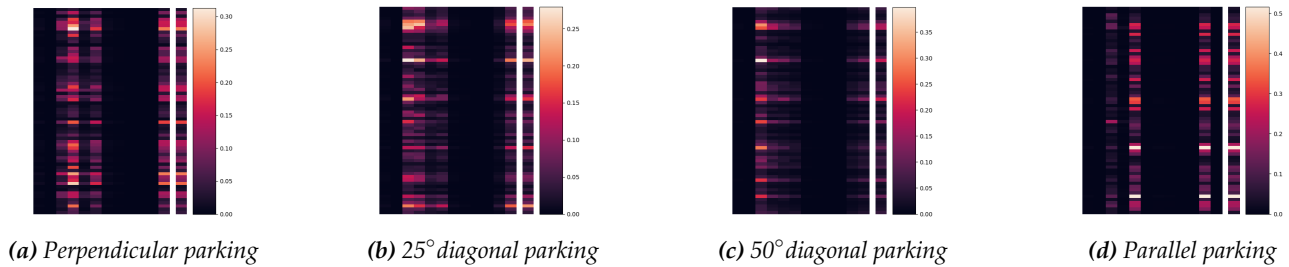

*(a) Perpendicular parking*  *(b) 25° diagonal parking*  *(c) 50° diagonal parking*  *(d) Parallel parking*

**Figure 3:** *Heatmap of EWC importances in the first layer after training on just a single scenario*

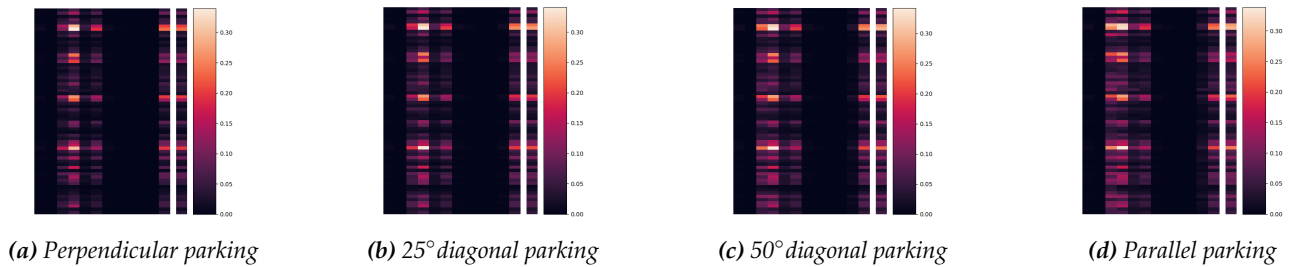

*(a) Perpendicular parking*  *(b) 25° diagonal parking*  *(c) 50° diagonal parking*  *(d) Parallel parking*

**Figure 4:** *Heatmap of EWC importances in the first layer after each scenario in sequential training*

One interpretation of this is transfer learning; the representations learned in perpendicular parking apply well to the other scenarios and so they are reused. Alternatively, existing latent representations may be favoured over training new representations as these already exist, even though they may not generalise well [13]. The second explanation is supported by an experiment in which the order of the scenarios was reversed, starting with parallel parking. In this case, none of the scenarios were learned successfully. This suggests the first scenario in the sequence introduces a bias that affects the learning of subsequent tasks. This also shows that the order of tasks has an effect on learning.

This raises the question of how to ensure that good representations are learned for different tasks. This is related to whether the capacity of neural networks allows for these different representations to be encoded in the neural network. It also poses the question of whether the sets of tasks or the order of tasks are compatible with each other and can be learned by the same network.

## 4 Research Agenda

To summarise, there are still significant open problems in CRL: finding suitable abstractions of the environment, oversensitivity to hyperparameters, catastrophic forgetting, and effective use of neural network capacity. Solving these problems is an important step to creating robust, generalisable CRL. Based on this work, we propose the following research questions for future work in advancing CRL:

1. **How can robust abstractions of the environment be constructed for all future tasks that the agent should learn, including unforeseen tasks?** Model-based CRL, where abstractions and rules are learned in a task-independent manner, may be an interesting approach to this. Specifically regarding reward models, generalisable reward functions and intrinsic rewards may also be interesting avenues of research.

2. **How can oversensitivity to hyperparameters be reduced in RL?** Possible approaches may include adaptive hyperparameters, defining algorithms that do not rely on hyperparameters, or enabling CLR agents to autotune hyperparameters at run time.

3. **How can catastrophic forgetting be prevented in complex domains such as CRL for cyber-physical systems?** The dependency on shared weights for different tasks appears to be significant barrier to solving this problem. This leads to the next question:

4. **How can the capacity of neural networks be used effectively so that a) suitable representations are learned and b) all future tasks (including unforeseen tasks) can be learned?** This may include aspects such as shaping which parts of the network are used for different tasks or functions, adding capacity, or adapting the structure of the network over time.

Overall, the challenges that these questions address can be traced back to fundamental limitations of neural networks: dependency on hyperparameters, limited capacity, and shared weights resulting in catastrophic forgetting. This calls into question whether neural networks are the best approach to CL.

A possible source of inspiration to answer these questions may be neuroscience. After all, biological systems have inspired many advances in artificial intelligence, including artificial neural networks, RL, and existing CL approaches such as EWC. As such, we believe that interdisciplinary research will play an important role in advancing CRL.